%% file: main.tex
\documentclass[10pt,twocolumn,letterpaper]{article}

\usepackage[review]{cvpr}      
\input{preamble}

\definecolor{cvprblue}{rgb}{0.21,0.49,0.74}
\usepackage[pagebackref,breaklinks,colorlinks,citecolor=cvprblue]{hyperref}

\def\paperID{20} 
\def\confName{CVPR}
\def\confYear{2024}

\title{EqvAfford: SE(3) Equivariance for Point-Level Affordance Learning}

\author{
  Yue Chen$^{1*}$\hspace{0.5cm}
  Chenrui Tie$^{2,1*}$\hspace{0.5cm}
  Ruihai Wu$^{1,3*}$\hspace{0.5cm}
  Hao Dong$^{1,3\dagger}$\hspace{0.5cm}\\
  \text{\textsuperscript{1}CFCS, School of CS, PKU}\hspace{0.5cm} 
  \text{\textsuperscript{2}School of EECS, PKU}\hspace{0.5cm} \\
  \text{\textsuperscript{3}National Key Laboratory for Multimedia Information Processing, School of CS, PKU}
}

\begin{document}

\maketitle

\renewcommand{\thefootnote}{\fnsymbol{footnote}} 
\footnotetext[1]{Equal contribution. Yue Chen and Chenrui Tie's order was determined by a coin flip. Ruihai Wu provided mentoring.}
\footnotetext[2]{Corresponding authors.}

\input{sec/Abs}
\input{sec/Introduction}
\input{sec/Related_work}
\input{sec/Prob}
\input{sec/Method}

\input{sec/Experiment}
\input{sec/Conclusion}
{
    \small
    \bibliographystyle{ieeenat_fullname}
    \bibliography{main}
}


\end{document}

%% file: preamble.tex
\usepackage[table,xcdraw,dvipsnames]{xcolor}


%% file: sec/Abs.tex
\begin{abstract}

Humans perceive and interact with the world with the awareness of equivariance,
facilitating us in manipulating different objects in diverse poses.
For robotic manipulation,
such equivariance also exists in many scenarios.
For example,
no matter what the pose of a drawer is (translation, rotation and tilt),
the manipulation strategy is consistent (grasp the handle and pull in a line).
While traditional models usually do not have the awareness of equivariance for robotic manipulation, 
which might result in more data for training and poor performance in novel object poses,
we propose our \textbf{EqvAfford} framework,
with novel designs to guarantee
the equivariance in point-level affordance learning for downstream robotic manipulation,
with great performance and generalization ability on representative tasks on objects in diverse poses.

\end{abstract}

%% file: sec/Introduction.tex
\section{Introduction}
\label{sec:introduction}
Symmetry is a fundamental aspect of our physical world, readily apparent in how we perceive and interact with the world. In this work, we focus on object manipulation tasks with certain spatial symmetry, where the manipulation strategy is equivariant to the object's 6D pose. This kind of spatial symmetry is called \textbf{SE(3) Equivariance},
where
\textbf{SE(3)} is a group consists of 3D translation and rotation.

We humans can naturally recognize and exploit SE(3) equivariance in object manipulation tasks. 
For example, the method of opening a drawer, pulling on its handle in a vertical direction, is intuitively understood to be effective regardless of the drawer's position, orientation, or tilt. 
However, the understanding of SE(3) equivariance poses a significant challenge for neural networks. 
Considering training a neural network to identify the grasp point on a drawer's point cloud,
if the point cloud is rotated and translated, the model's output could be highly different,
while the manipulation point and strategy should be consistent.
This case underscores the limitations of data-driven approaches that lack an understanding of the physical world, leading to weak generalization across varied initial conditions.
In this work, we develop a perception system designed to deduce potential affordance, which includes both the location of interaction and the manner of manipulation, by leveraging SE(3) equivariance to bolster the model's generalization ability.

\begin{figure}
\centering
\includegraphics[width=0.47\textwidth]{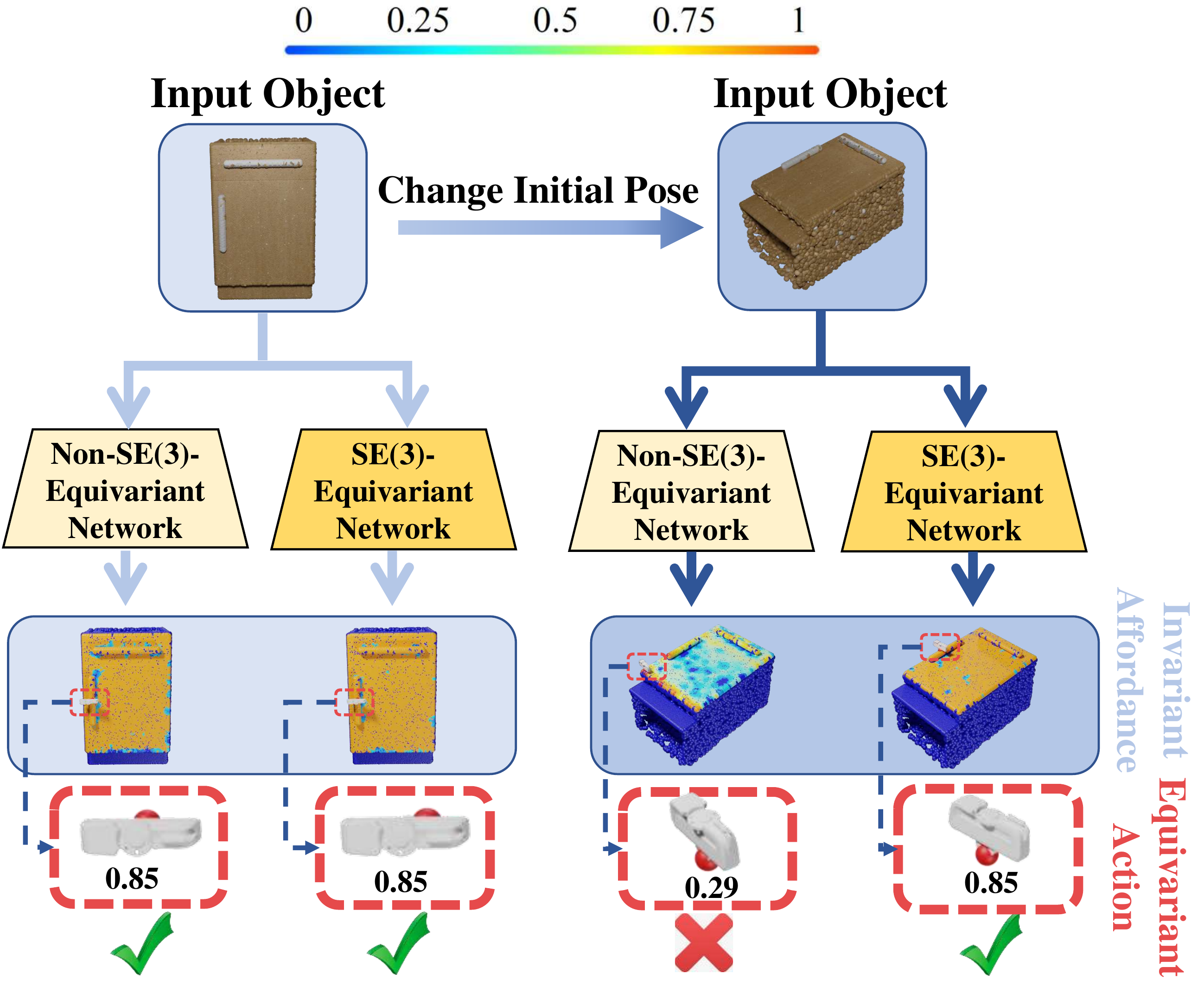}
\caption{For each point on a
3D object, we predict an actionability (affordance) score and an interaction orientation to manipulate the object. By leveraging SE(3) equivariance, our method shows consistency and generalizes to different poses of the object.}
\vspace{-5mm}
\label{fig1poses}
\end{figure}

Object manipulation stands as a pivotal area of research in robotics, where a key challenge lies in identifying the optimal interaction points (\emph{i.e.}, affordable point that indicates where to interact with) and executing appropriate actions (how to interact). We discover that a large portion of object manipulation tasks exhibit SE(3) equivariance. Notably, for tasks where the interaction is independent of the object's absolute spatial pose, leveraging equivariance can significantly help.
 In the case of the drawer opening task, the optimal strategy remains consistent regardless of the drawer's rotation: grasp the handle and exert a vertical pull outwards. 
 Therefore,
we propose a framework that leverages SE(3) equivariance to predict per-point affordance and downstream manipulation strategy on each point, with theoretical guarantees and great performance.

In summary, our main contributions are:
\begin{itemize}
    \item We propose to leverage the inherent equivariance for point-level affordance learning and downstream robotic manipulation tasks;
    \item We propose a framework with novel designs and multiple modules that can infer affordance for object manipulation with the theoretical guarantee of equivariance;
    \item Experiments show the superiority of our framework on manipulating objects with diverse and novel object poses.
\end{itemize}

%% file: sec/Related_work.tex
\section{Related Work}
\label{sec:related_works}

\subsection{Learning Affordance for Robotic Manipulation}
For manipulation, it's crucial to find suitable interaction points. Gibson proposed affordance – interaction opportunity. 
They leveraged observations as demonstrations, like videos showing agents interacting with scenarios, and trained models to learn what and how to interact. This kind of method has been used to learn possible contact locations\cite{brahmbhatt2019contactdb} or grasp patterns\cite{hamer2010object}.
However, using passive observations to learn possible interactions suffers from distribution shifts. \cite{mo2021where2act} proposed an approach learning from interactions to obtain more informative samples for articulated object manipulation\cite{jain2021screwnet,hu2017learning}.
Recent works expanded affordance to more diverse scenarios and tasks, such as articulated object manipulation~\cite{yuan2024generalflowfoundationaffordance,wang2024rpmartrobustperceptionmanipulation,geng2024sagebridgingsemanticactionable}, deformable object manipulation~\cite{ luunigarment, CVPR, wu2023learning}, clutter manipulation~\cite{li2024broadcasting}, bimanual manipulation~\cite{zhao2022dualafford}, and zero-shot generalization to novel shapes~\cite{ju2024robo, kuang2024ramretrievalbasedaffordancetransfer}.

\subsection{Leveraging SE(3) Equivariant Representations for Robot Manipulation}

Many works focused on building SE(3) equivariant representations from 3D inputs\cite{deng2021vector,fuchs2020se,lei2023efem,ryu2022equivariant,weng2023neural,seo2023robot,yang2024equibotsim3equivariantdiffusionpolicy,Liu_2023_ICCV,scarpellini2024diffassemble}.
to enable robust generalizations to out-of-distribution inputs by ensuring inherent equivariance. 
This allows the model to automatically generalize to inputs at novel 6D poses, without implementing strong data augmentation.
Prior works have applied equivariant representations on diverse robot manipulation tasks~\cite{xue2023useek,simeonov2022neural,simeonov2023se,wu2023leveraging,yang2023equivact,wang2024equivariantdiffusionpolicy,hu2024orbitgraspse3equivariantgrasplearning,huang2022edgegraspnetworkgraphbased,gao2024riemannnearrealtimese3equivariant,Ryu_2024_CVPR}. 
This work studies leveraging SE(3) equivariance for point-level affordance learning and thus complete articulated object manipulation.

%% file: sec/Prob.tex
\section{Problem Formulation}
\label{sec:prob}

We tackle 3D articulated object manipulation tasks. Given a point cloud $x$ of a 3D articulated object ($e.g.$ a cabinet with drawers and doors), our model predicts \textit{where} and \textit{how} to interact with the object.
We define 6 types of short-term primitive actions like pushing or pulling, which are parameterized by the 6D pose of the robot gripper.
For each primitive action, we predict the following features for each visible point $p$ of the 3D articulated object.
\begin{itemize}
    \item An affordances score $a_p\in [0,1]$, measuring how likely the point $p$ should be interacted with in the specific task.
    \item A 6D pose $R_{p}\in SO(3)$ to interact with the point $p$. 
    \item A success likelihood score $s_p^R$ for an action proposal $R_{p}$.
\end{itemize}

The affordance score $a_p$ should be invariant while the proposed 6D posed $R_p$ should be equivariant for any SE(3) transformation applied on the object. For example, ``pulling drawer'' task, if the cabinet is rotated or translated, the $a_p$, which means how likely the point $p$ will be interacted with, should remain invariant. Because no matter how the cabinet is placed, the correct way to open the drawer is always to interact with the handle. And the 6D interaction pose needs to change corresponding to the pose change of the cabinet.

%% file: sec/Method.tex
\section{Method}
\label{sec:method}

\begin{figure}
\centering
\includegraphics[width=0.47\textwidth]{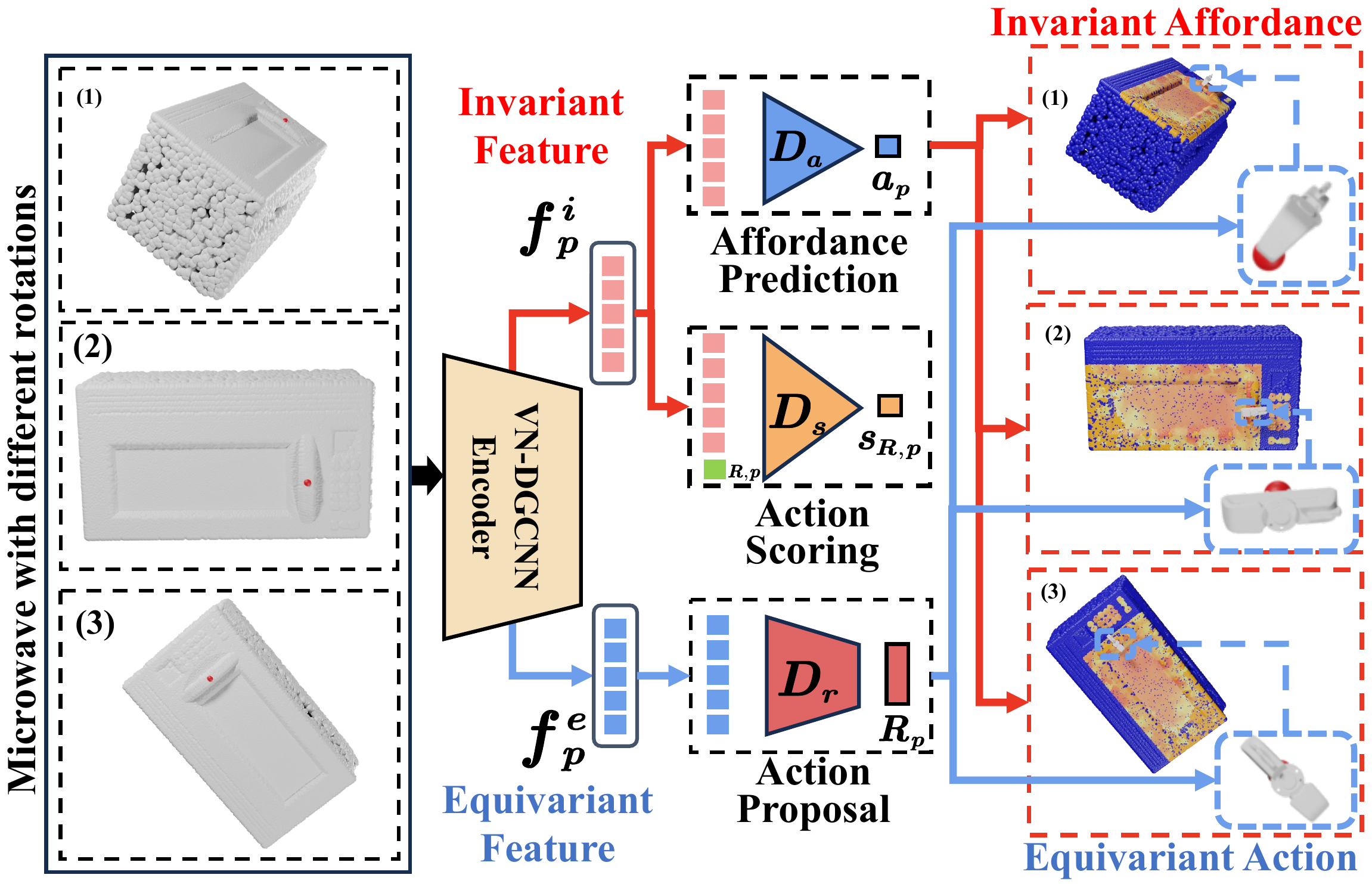}
\caption{\textbf{Overview of our proposed framework.}
Taking as input a point cloud of the object, our framework first outputs a per-point SE(3) invariant feature $f_p^i$ and SE(3) equivariant feature $f_p^e$. The invariant $f_p^i$ results in the affordance map invariant to object rotations,
while the equivariant feature $f_p^e$ results in the manipulation actions equivariant to object rotations. 
}
\vspace{-5mm}
\label{fig2}
\end{figure}

\subsection{Network Designs}

As shown in Fig.\ref{fig2}, our framework consists of four modules: \textbf{VN-DGCNN Encoder}, \textbf{Affordance Prediction Module}, \textbf{Action Proposal Module}, and \textbf{Action Scoring Module}.

The \textbf{VN-DGCNN Encoder} is the first encoder of objects with invariant and equivariant features.

During inference, the \textbf{Affordance Prediction Module} predicts the point-level affordance score (whether an point is actionable for the task) and thus selects the manipulation point with the highest affordance score,
the \textbf{Action Proposal Module} proposes many candidate actions of the gripper,
and the \textbf{Action Scoring Module} rates the proposed actions and then selects the best action for manipulation.

During training,
we first collect diverse interactions with corresponding interaction results to train the \textbf{Action Scoring Module} and \textbf{Action Proposal Module},
and the \textbf{Affordance Prediction Module} is then trained to aggregate the average scores rated by \textbf{Action Scoring Module} of actions proposed by \textbf{Action Proposal Module}.

\textbf{SE(3) Equivariant Encoder.}
For each point of the input point cloud, we extract a SE(3) invariant feature $f_p^i$ and a SE(3) equivariant feature $f_p^e$.  Here we use the VN-DGCNN segment network \cite{deng2021vector} to extract the features. 
Denote the number of points in the input point cloud as $N$, length of feature as $d$.
As \cite{deng2021vector} use vectors to represent neurons, $f^i_p,f^e_p\in \mathbb{R}^{d\times3}$.
For each point $p\in \mathbb{R}^3$ in the input point cloud $x\in \mathbb{R}^{N\times 3}$, we have
\begin{align}
\mathcal{E}_i(p) &= f_p^i \label{eq:Ei} \\
\mathcal{E}_e(p) &= f_p^e \label{eq:Ee}
\end{align}

When point cloud $x$ transformed by $T$ $\in SE(3)$, we have
\begin{align}
\mathcal{E}_i(T(p)) &= f_p^i \label{eq:EiTp} \\
\mathcal{E}_e(T(p)) &= T(f_p^e) \label{eq:EeTp}
\end{align}

\textbf{Affordance Prediction Module.} Our affordance prediction module, similar to Where2Act\cite{mo2021where2act}, is a MLP. It takes the SE(3) invariant feature $f_p^i$ of each point and outputs a per-point affordance $a_p\in [0,1]$. As the input is invariant to any translation and rotation, the affordance $a_p$ is SE(3) invariant. 
This means regardless of the position and orientation of the input object, we should always prefer to interact with the same part of the object (such as handle when opening a drawer).
Formally,\begin{equation} 
a_p = D_a(f_p^i)
\end{equation}

\textbf{Action Pose Proposal Module.}
For each point $p$, it takes the SE(3) equivariant feature $f_p^e$ as input. We employ VN-DGCNN classification network\cite{deng2021vector} to implement this module, which is equivariant to any SE(3) transformation. The network $D_p$ predicts a gripper end-effector pose as a rotation matrix, which is equivariant to the input point cloud. 
\begin{equation} 
R_p = D_r(f_p^e)
\end{equation} 

\textbf{Action Scoring Module.}
For a proposed interaction orientation $R$ at point $p$, we use a MLP to evaluate a success likelihood $s_{R,p}\in [0,1]$ for this interaction. This score can be used to filter low-rated proposals or select the proposals that are more likely to succeed.
Given a invariant feature $f_p^i$ of point $p$, and a proposed interaction orientation $R$, this module outputs a scalar$s_{R,p}\in [0,1]$. Namely,
\begin{equation} 
s_{R,p} = D_s(f_p^i,R)
\end{equation} 
At test time, the interaction is positive when $s_{R,p}>0.5$.
Due to the page limit,
more details of point-level affordance designs can be found in ~\cite{mo2021where2act, wu2021vat}.

\subsection{Data Collection}
Following \cite{mo2021where2act}, we adopt a learning-from-interaction method to collect training data. In the SAPIEN\cite{xiang2020sapien} simulator, we place a random articulated object at the center of the scene and use a flying parallel gripper to interact with the object on a specific point $p$ and orientation $R$. We tackle six types of primitive action and pre-program a short trajectory for each primitive action. When $(p,R)$ is specified, we can try different pre-programmed short trajectories to interact with the object.

We use both online and offline methods to collect training data. In offline collection, we randomly choose an object and load it into the simulator. We then sample an interaction point $p$ and an interaction orientation $R$ and roll out to see whether the gripper can complete the task.

As offline data collection is inefficient, we also add online adaptive data collection. When training the action scoring module $D_s$ with data pair $(p,R)$, we infer the score predictions $s_{R,p}$ for all points on the articulated object. Then we sample a point $p^*$ with highest score to conduct an addition interaction with point $p^*$ and orientation $R$. 

\subsection{Training and Losses}
Our loss function is a combination of following three terms.

\textbf{Action Scoring Loss.} For a batch of data points $(p,R,r)$, where $r$ is the ground-truth interaction result($1$ for positive, $0$ for negative). We train the action scoring module $D_s$ with the standard binary cross-entropy loss.

\textbf{Action Proposal Loss.} For a batch of data points $(p,R,r)$, we train the action pose proposal module $D_p$ on positive data pieces by calculating the geodesic distance between the predicted orientation $R_p$ with the ground truth orientation $R$.

\textbf{Affordance Prediction Loss.} We technically train the affordance prediction module $D_a$ by supervising the expected success rate when executing a random orientation proposed by $D_r$. Instead of rolling out this proposal in the simulator, we directly use the action scoring module $D_s$ to evaluate whether an interaction is successful.

%% file: sec/Experiment.tex
\section{Experiment}
\label{sec: experiment}
We use SAPIEN~\cite{xiang2020sapien} simulator with realistic physical engine to evaluate our framework, and compare with Where2Act~\cite{mo2021where2act}, with its same setting for fair comparison.

We select 15 object categories in the PartNet-Mobility dataset~\cite{xiang2020sapien} to conduct our experiments and report the quantitative results on four commonly seen tasks: pushing and pulling doors and drawers.
We train our method and baseline over training shapes and report performance over test shapes from the training categories and shapes from the test categories to test the generalization capabilities over novel shapes and unseen categories

\begin{figure*}[ht]
\centering
\includegraphics[width=1\textwidth]{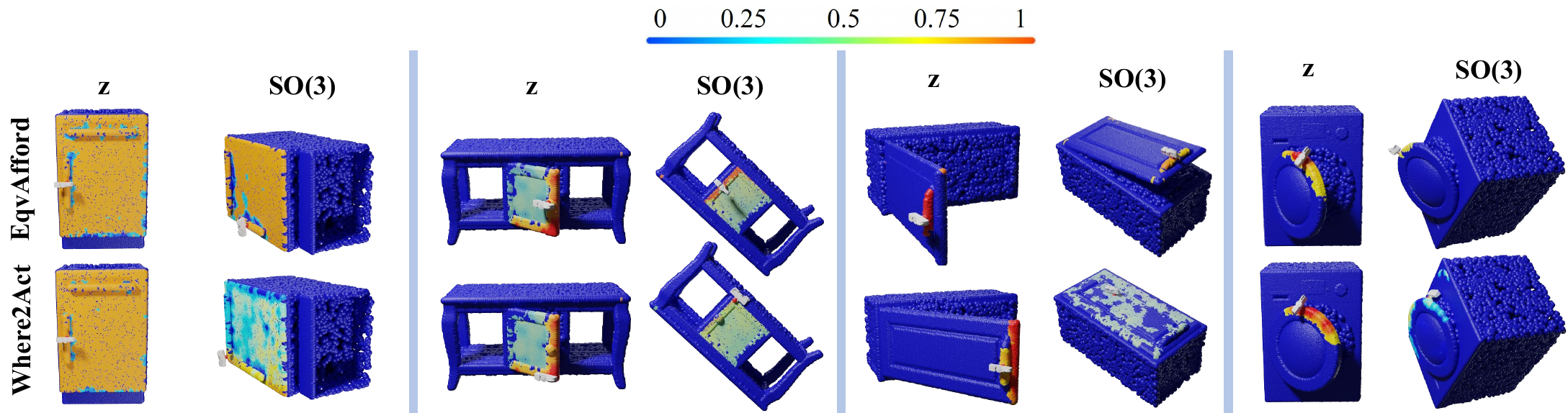}
\caption{We visualize the action scoring and action proposal predictions over the movable parts. We show the qualitative results of our method and baseline on two experiment settings across four objects. Our method is robust to rotation of the input point cloud.} 
\label{fig3}
\end{figure*}

\begin{table*}[ht]
\centering
\begin{tabular}{@{}lccccp{1cm}@{}}
\toprule
               & \multicolumn{2}{c}{Where2Act}                         & \multicolumn{2}{c}{EqvAfford}                                \\ \midrule
               & z             & \cellcolor[HTML]{F8D2CF}SO(3)         & z             & \cellcolor[HTML]{6D9EEB}SO(3)                  \\ \midrule
Pushing Door   & 86.55 / 80.42 & \cellcolor[HTML]{F8D2CF}56.74 / 53.44 & 86.49 / 81.03 & \cellcolor[HTML]{6D9EEB}\textbf{86.79 / 81.55} \\
Pulling Door   & 64.29 / 59.87 & \cellcolor[HTML]{F8D2CF}56.39 / 51.82 & 63.81 / 58.55 & \cellcolor[HTML]{6D9EEB}\textbf{65.70 / 62.65} \\
Pushing Drawer & 86.49 / 86.99 & \cellcolor[HTML]{F8D2CF}58.31 / 48.49 & 87.89 / 87.09 & \cellcolor[HTML]{6D9EEB}\textbf{88.81 / 87.76} \\
Pulling Drawer & 67.05 / 60.35 & \cellcolor[HTML]{F8D2CF}58.83 / 52.15 & 67.74 / 61.01 & \cellcolor[HTML]{6D9EEB}\textbf{68.22 / 61.96} \\ \hline
\end{tabular}
\caption{Quantitative evaluation of the learned point-level affordance using F1-score. The numbers correspond to test shapes from both the training categories (before slash) and the test categories (after slash). 
Higher numbers indicates better performance.}
\label{tb:score}
\end{table*}

\begin{table*}[!h]
\centering
\begin{tabular}{@{}lccccp{1cm}@{}}
\toprule
               & \multicolumn{2}{c}{Where2Act}                         & \multicolumn{2}{c}{EqvAfford}                                \\ \midrule
               & z             & \cellcolor[HTML]{F8D2CF}SO(3)         & z             & \cellcolor[HTML]{6D9EEB}SO(3)                  \\ \midrule
Pushing Door   & 83.91 / 81.49 & \cellcolor[HTML]{F8D2CF}38.44 / 37.78 & 83.84 / 82.80 & \cellcolor[HTML]{6D9EEB}\textbf{85.51 / 83.96} \\
Pulling Door   & 62.76 / 57.28 & \cellcolor[HTML]{F8D2CF}11.95 / 11.77 & 62.06 / 56.98 & \cellcolor[HTML]{6D9EEB}\textbf{62.78 / 59.23} \\
Pushing Drawer & 78.39 / 74.89 & \cellcolor[HTML]{F8D2CF}35.37 / 29.45 & 79.58 / 76.02 & \cellcolor[HTML]{6D9EEB}\textbf{80.16 / 77.10} \\
Pulling Drawer & 68.26 / 64.45 & \cellcolor[HTML]{F8D2CF}12.65 / 12.26 & 68.88 / 65.96 & \cellcolor[HTML]{6D9EEB}\textbf{69.16 / 67.10} \\ \bottomrule
\end{tabular}
\caption{We report task success rates 
of manipulating articulated
objects.
We evaluate our method and baseline in four tasks. 
The numbers correspond to test shapes from both the training categories (before slash) and the test categories (after slash).
}
\label{tb:success}
\end{table*}

\textbf{Results and Analysis.}
To demonstrate our method keeps the inherent equivariance, we designed two test settings, \textbf{$z$} and \textbf{$SO(3)$}. Where \textbf{$z$} stands for placing objects on an aligned orientation and generates novel  with rotations only along the z-axis, and \textbf{$SO(3)$} for arbitrary rotations.
Table ~\ref{tb:score} and ~\ref{tb:success} provide quantitative analysis comparing our method to the Where2Act baseline, revealing that \textbf{EqvAfford} stands out as the most effective in both affordance learning and downstream robotic manipulation.
In table~\ref{tb:score}, Our method adeptly captures invariant geometric features, outperforming Where2Act, whose features fluctuate with rotational poses. 
And we visualize the predicted action scores in Fig \ref{fig3}, where we can see that given Objects with different poses, our method learns to extract invariant geometric features that are action-equivariant. In contrast, the baseline models fail to achieve a comparable level of performance. 
Table \ref{tb:success} further underscores our superiority by showcasing the higher task success rates in manipulating articulated objects.
At Last, our model exhibits enhanced generalization, effectively adapting to unseen novel object categories beyond the baseline models. 
\vspace{-0.8cm} 

%% file: sec/Conclusion.tex
\vspace{-3mm}
\section{Conclusion}
\vspace{-3mm}
\label{sec:conclusion}
We propose a novel framework, \textbf{EqvAfford}, leveraging SE(3) equivariance for affordance learning and downstream robotic manipulation with novel designs to theoretically guarantee equivariance.
Predicting per-point \textbf{SE(3) Invariant} affordance and \textbf{ Equivariant} interaction orientation, our method generalizes well to diverse object poses. Experiments on affordance learning and robotic manipulation showcases our method qualitatively and quantitatively.